\documentclass[10pt,twocolumn,letterpaper]{article}

\usepackage{graphicx}
\usepackage{amsmath}
\usepackage{amssymb}
\usepackage{booktabs}
\usepackage{svg}
\usepackage{mathtools}
\usepackage{xcolor}
\usepackage{comment}
\usepackage{adjustbox}

\newcommand{\etalc}[2]{#1 \textit{et al.} \cite{#2}}
\newcommand{\cens}[1]{xxx}

\usepackage[pagebackref,breaklinks,colorlinks]{hyperref}

\usepackage[capitalize]{cleveref}
\crefname{section}{Sec.}{Secs.}
\Crefname{section}{Section}{Sections}
\Crefname{table}{Table}{Tables}
\crefname{table}{Tab.}{Tabs.}

\begin{document}

%\title{A low-parameter camera pipeline for night photography rendering}
%\title{SCIALLA: a shallow camera pipeline for night photography rendering}
\title{Shallow camera pipeline for night photography rendering}

\author{Simone Zini*
\and
Claudio Rota
\and
Marco Buzzelli
\and
Simone Bianco
\and
Raimondo Schettini \\ \\
University of Milano - Bicocca\\
Milano, Italy\\
{\tt\small s.zini1@campus.unimib.it}
}
\maketitle

%%%%%%%%% ABSTRACT
\begin{abstract}
We introduce a camera pipeline for rendering visually pleasing photographs in low light conditions, as part of the NTIRE2022 Night Photography Rendering challenge.
Given the nature of the task, where the objective is verbally defined by an expert photographer instead of relying on explicit ground truth images, we design an handcrafted solution, characterized by a shallow structure and by a low parameter count.
Our pipeline exploits a local light enhancer as a form of high dynamic range correction, followed by a global adjustment of the image histogram to prevent washed-out results.
We proportionally apply image denoising to darker regions, where it is more easily perceived, without losing details on brighter regions.
The solution reached the fifth place in the competition, with a preference vote count comparable to those of other entries, based on deep convolutional neural networks.
Code is available at www.github.com/AvailableAfterAcceptance
%https://github.com/TheZino/shallow-camera-night-photo
\end{abstract}

%%%%%%%%% BODY TEXT
\section{Introduction}
\label{sec:intro}

Images in low light conditions are usually shot for surveillance, or for collection of personal memories.
The NTIRE2022 Night Photography Rendering challenge defines a focus on the latter, explicitly aiming at rendering procedure that produces ``visually pleasing photographs''.
In this work we present our solution to the challenge.

A digital image is generated through a camera-specific pipeline, that processes the RAW sensor data in order to produce a final RGB output.
These imaging pipelines are typically composed of fixed steps, such as demosaicing, as well as adaptive steps that depend on the image content, such as automatic white balance (AWB).
With respect to daytime photography, night photography presents a specific set of challenges, that should be taken into account in the development of a camera pipeline, in order to strive for improved visual pleasantness.
Most notably, sensor noise is typically more prominent in low-light conditions, due to the need for increased sensor sensitivity.
Furthermore, white balance mechanisms of human vision are known to work differently between daytime and night time, with relative fewer studies covering the latter case and its computational counterpart.

% Rai:
At a time when the resolution of most machine learning problems is delegated to the direct or indirect use of neural networks, we address the challenge with a ``traditional'' processing pipeline, whose main modules are designed on the basis of our knowledge of the mechanisms of human vision, and on the basis of our knowledge of the main limitations of traditional imaging devices. The, few, parameters of the different modules are heuristically set by the authors according to their personal preferences \cite{bianco2020personalized}.
This low parametric dependency means that our solution is flexible, as it can be potentially tuned to match individual users preferences and to different sensors.
We consider this characteristic to be particularly relevant in the context of this challenge, where no ground truth nor evaluation metric is provided, as entries are evaluated solely based on human judgment.
%  i parametri scelti a mano e non presi da altri paper sono:
% - i vari sigma per il gaussian blur
% - i sigma per il denoising
% - il percentile del black stretch
% - gamma 1/1.4 global
% - mask noise blending
% altri possibili parametri, come quelli del GI, sono presi dal paper o comunque dalla loro implementazione

\section{Proposed method}
\label{sec:method}

\begin{figure*}
    \centering
    \adjustbox{width=\textwidth}{
    \includegraphics{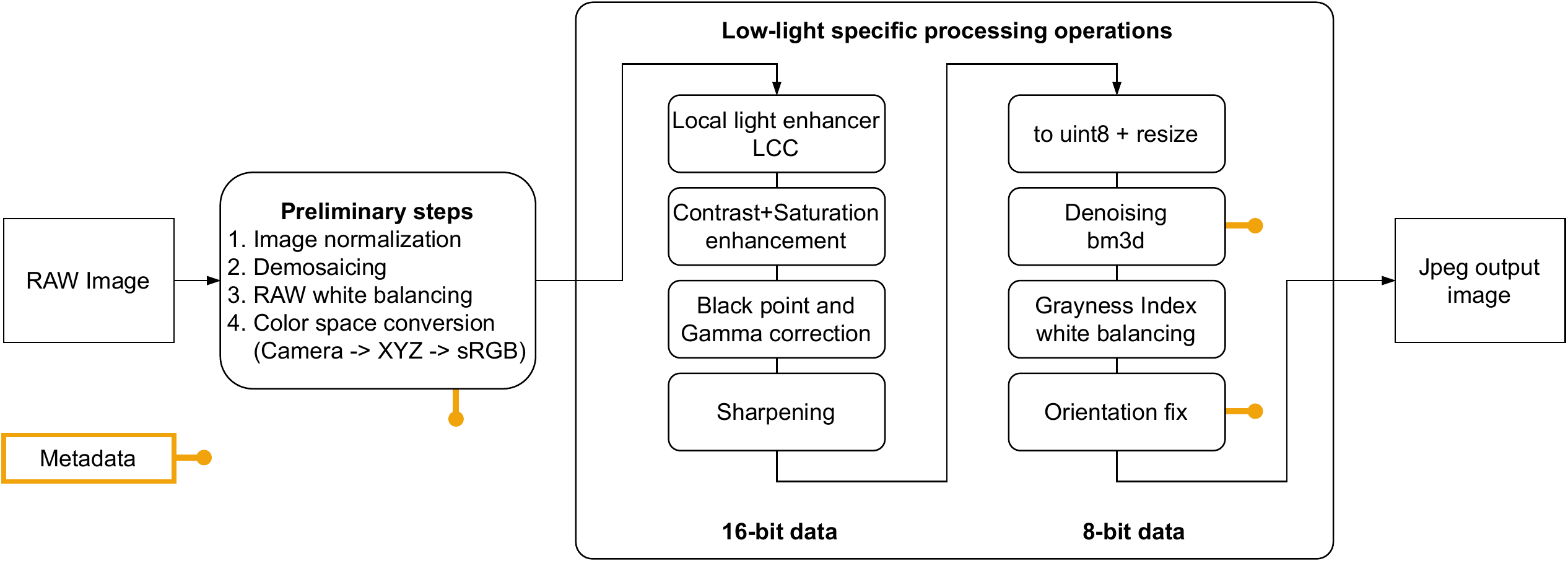}
    }
    \caption{Overview of the complete proposed pipeline. The entire pipeline can be divided in two parts: preliminary data preparation steps and low-light processing steps. Metadata extra information is exploited in the steps marked with the orange dot.}
    \label{fig:pipeline}
\end{figure*}

The scheme of the proposed solution is depicted in Figure~\ref{fig:pipeline}. Our pipeline can be divided into two parts: the preliminary steps, which are the basic stages of a typical camera processing pipeline, and the low-light specific part, which instead contains steps to specifically handle night images.
We refer multiple times within our pipeline to image metadata, which are indicated in the following using \textit{italic text}.

\subsection{Preliminary steps}
The first part of our pipeline is made of four steps working in the RAW domain.
The first step is image normalization:
the \textit{black\_level} as provided in the image metadata is subtracted, and the image values are rescaled so that the \textit{white\_level} is set to one.
The demosaicing operation converts the single-channel RAW image into the three-channel RGB image using the appropriate color filter array pattern (\textit{cfa\_array\_pattern}).
Then, a preliminary automatic white balance step is performed using the Gray World algorithm \cite{gw}, in order to provide a first approximate correction of the image cast.
Finally, a color transformation step converts the image from the camera-specific color space to XYZ (obtained as the inverse of \textit{color\_matrix\_1}) and finally to the sRGB color space.

\subsection{Low-light specific processing operations}
The second part of our pipeline has been specifically designed to handle images taken by night in low-light conditions. 

The first step of this second part is the use of the Local Contrast Correction (LCC) algorithm by Moroney \cite{mcc}. Here the local correction is performed on the Y channel of the YCbCr color space using a pixel-wise gamma correction, whose values are determined using a mask $M$ obtained by blurring the luminance channel Y with a Gaussian filter in order to brighten dark areas and to not clip pixels that are already bright. The corrected $\hat{Y}$ channel image is obtained as 
\begin{equation}\label{eq:lcc}
\hat{Y} = Y^{\gamma^{\frac{0.5 - (1 - M)}{0.5}}},
\end{equation}
where $M$ is computed as previously described, and $\gamma$ is the value of the exponent for gamma correction. According to \etalc{Schettini}{lccschett}, we computed $\gamma$ as:
\begin{equation}
    \gamma = 
    \begin{dcases}
        \frac{\ln (0.5)}{\ln (\bar{Y})} & \text{if} \: \bar{Y} \geq 0.5 \\
        \frac{\ln (\bar{Y})}{\ln 0.5} & \text{otherwise}
    \end{dcases},
\end{equation}
where $\bar{Y}$ is the average value of the Y channel. Since $1 - M$ inverts the computed mask, bright areas are darkened by a gamma value lower than 1, and dark areas are brightened by a gamma value greater than 1.

The application of LCC tends to reduce the overall contrast and saturation, as noted by \etalc{Schettini}{lccschett}. Therefore, as subsequent steps, we perform contrast and saturation enhancement.

The contrast enhancement step adaptively stretches and clips the image histogram based on how the distribution of dark pixels changes before and after the contrast correction of LCC.
Every histogram computed has 256 bins.
The histogram range used for stretching and clipping is defined as follows: let any given pixel be ``dark'' if, in the YCbCr color space, its Y value is lower than 0.14 and its chroma radius is lower than 0.07, computed as:
\begin{equation}
    CR = \frac{(Cb - 0.5) \times 2 + (Cr - 0.5) \times 2}{2},
\end{equation}
The lower range for histogram stretching is defined by the number of dark pixels after the application of LCC.
If there is at least one pixel, the lower range is given by the difference of the bins corresponding to 30\% of dark pixels in the cumulative histogram of $\hat{Y}$ and $Y$, which represent the output and input of LCC in Equation \ref{eq:lcc}.
If there are no dark pixels, the lower range corresponds to the 2nd percentile value of the $\hat{Y}$ histogram.
Concerning the ``bright'' pixels, the upper range for histogram stretching always corresponds to the 98th percentile value of the $\hat{Y}$ histogram.
For both ranges, the maximum number of bins to clip is 50. Using the determined range, the image histogram is stretched and the histogram bins that fall outside are clipped.

For the saturation enhancement step, we correct each RGB channel as suggested by \etalc{Sakaue}{saturationfix}:
\begin{equation}
    \hat{C} = 0.5 \times \frac{\hat{Y}}{Y} \times (C + Y) + C - Y,
\end{equation}
where $C$ stands for each RGB channel, $\hat{C}$ is the corresponding output channel, $\hat{Y}$ and $Y$ are the output and input Y channels used in Equation~\ref{eq:lcc}.

After contrast and saturation enhancement, a black point correction step is performed in order to restore the natural aesthetics of night images, since LCC adjusts local statistics but produces an overall washed-out result. This operation is performed by clipping to zero all pixels below the 20th percentile value of the value channel V in the HSV color space. After this operation, a global gamma correction is performed with gamma value set to $\frac{1}{1.4}$, followed by a sharpening operation using unsharp masking.

The image is then converted to 8 bit encoding, resized to match the %\textit{out\_landscape\_width} and \textit{out\_landscape\_height} metadata,
predefined output size (imposed by the challenge to be $1300\times866$ for landscape orientation and $866\times1300$ for portrait one), and processed with the Block-Matching and 3-D Filtering (BM3D) denoising algorithm \cite{bm3d} to remove noise introduced by the poor light conditions typical of night scenes.
Here the \textit{noise\_profile} value from the image metadata is used to determine the strength of the denoising operation, which is controlled by BM3D through a parameter $\sigma$ that encodes an estimate of the noise standard deviation, used internally to control the parameters of the method.
According to the distribution of the \textit{noise\_profile} values in the training data, we defined three classes representing different noise intensities, and we empirically assigned a $\sigma$ value (0.2, 0.6 and 0.8) to each class.
Since noise is more visible in dark regions rather than in bright regions and BM3D removes part of the high frequency information, 
%the final denoised image is obtained by blending the image denoised using BM3D with the original noisy image, so that part of the details in brighter areas can be preserved. We
we performed a blending operation in RGB using a mask generated by blurring the luminance channel Y of the original noisy image in the YCbCr color space with a Gaussian filter. The final denoised image $\hat{D}$ is computed as
\begin{equation}\label{eq:blend}
    \hat{D} = I_{BM3D} \times (1 - mask \times u) + I \times (mask \times u), 
\end{equation}
where $I_{BM3D}$ is the image denoised with BM3D, $I$ is the original noisy image and the $u$ parameter, empirically set to 0.6, controls the denoising effect in bright areas.

A second automatic white balance step is performed, this time on non-linear processed RGB data, in order to reduce color casts in those scenarios where the initial Gray World approach may have failed.
Here the Grayness Index (GI) algorithm \cite{GI} is used. GI is very sensible to noise, hence we estimated the image illuminant on the image $I_{BM3D}$, then we normalized it by its maximum value and applied it to the image $\hat{D}$ obtained after the blending operation in Equation \ref{eq:blend}.

The image is then rotated in relation to the \textit{orientation} information stored in the metadata and finally saved as JPEG image at quality 100.

\begin{comment}
% non so quanto possa essere utile questa parte visto che abbiamo specificato tutto nel metodo, da valutare
\section{Implementation details}
\label{sec:implementationdetails}
Our pipeline is developed in Python and publicly available \textbf{here}. The preliminary steps for RAW processing correspond to the original pipeline provided by the challenge organizers, from which we removed the denoising step.
\end{comment}

\section{Experiments}
\label{sec:experiments}

%* setup (dataset)\\
%* evaluation protocol\\
%* results (classifica e commenti fotografi)\\

\subsection{Dataset}
The NTIRE2022 Night Photography Rendering challenge provided 250 RAW-RGB images of night scenes captured using the same sensor type and encoded in 16-bit PNG files. Each RAW image has a resolution of $3646\times5202$ pixels. Image metadata are also available in JSON format. Due to the nature of the challenge, ground truth images are not available.

According to the challenge organization, 50 images are provided as train set, 50 as the first validation set, 50 as the second validation set and the remaining 100 as the final validation set (among these 100, only 50 were selected for the final evaluation). Since our solution does not need a training procedure, we used the train set to empirically select the few parameters required by our pipeline, and used all validation sets to validate the results. 

\subsection{Evaluation protocol}
The challenge results have been evaluated according to two different protocols, each resulting in a different leaderboard. The two protocols are described in the following.

\subsubsection{People choice protocol}

The evaluation of the results has been done through Mean Opinion Score (MOS) using the Yandex Toloka platform, with a pool of observers from Eastern Europe and Russia to perform the image ranking. As pointed out by the challenge organizers, there may be a cultural bias in terms of the preferred image aesthetics due to the geographical distribution of the observers.

For this evaluation no guidelines have been provided, so we mainly followed those provided by the professional photographer, described in the next subsection.
%\TODO{The final evaluation, whose results are reported in Table~\ref{tab:mos}, refers to a subset of 50 images taken from the final validation set provided.}

\subsubsection{Professional photographer choice protocol}
\begin{figure}[t]
    \centering
    \includegraphics[width=\columnwidth]{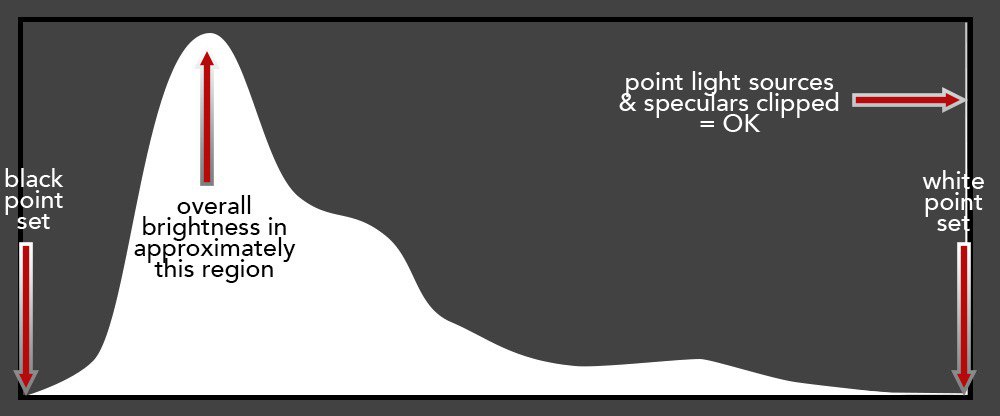}
    \caption{General image histogram distribution expected by the professional photographer. The overall histogram should be biased to the left and occupy the full dynamic range. White should correspond to point light sources and speculars.}
    \label{fig:professionalphotographer}
\end{figure}

\begin{figure*}[t]
    \centering
    \adjustbox{width=\textwidth}{
    \includegraphics{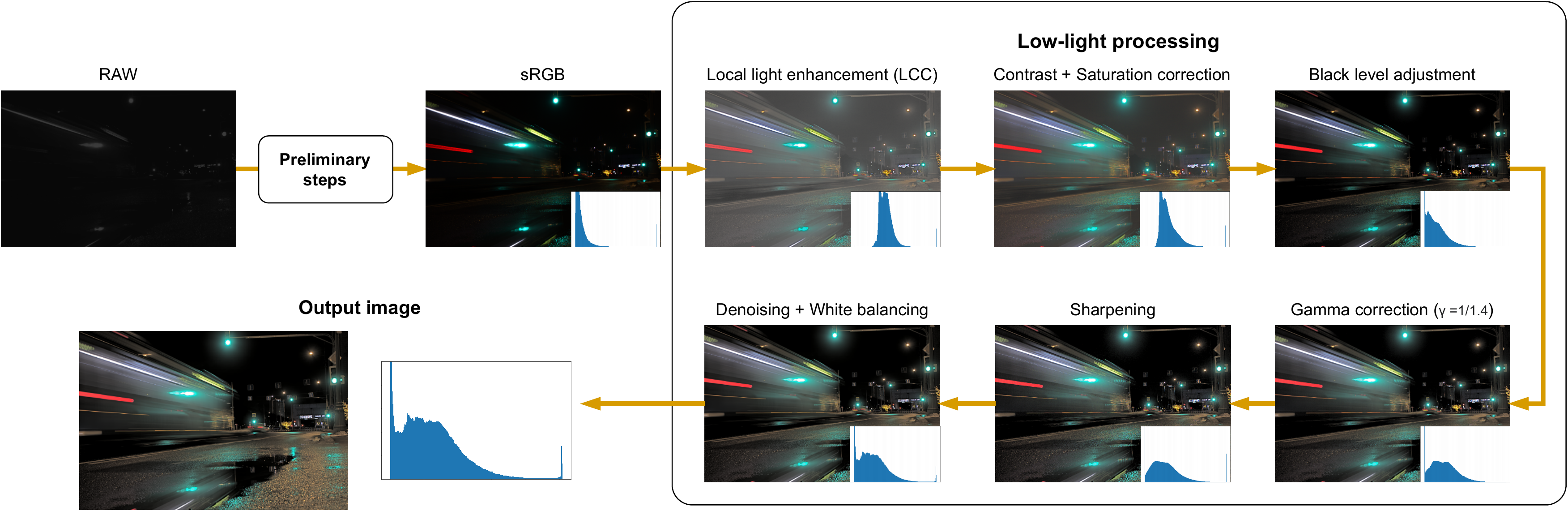}
    }
    \caption{Step-by-step results of the proposed pipeline. Along with images, we also reported histograms to show how the global pixel distribution changes.}
    \label{fig:hist_pipe}
\end{figure*}

A professional photographer was involved throughout the challenge, providing tips and personal expectations that guided the development of our pipeline. 
The properties of correctly rendered images have been classified in two main groups, under the names of \textit{(a) image integrity} and \textit{(b) interpretation}.

Group \textit{(a)} has mainly to do with artifacts, focusing on objective aspects of the images. Examples of considered artifacts are: the camera noise, light source and specular artifacts, including sharp clipping edges surrounding these, and colour banding.

Group \textit{(b)} instead covers more subjective aspects related to the photographer experience and the photography industry field.
Here aspects like the average brightness, tonal range from black point to white point, global contrast, local contrast, dominant hue and saturation are considered.
An histogram showing the expected image light distribution provided by the professional photographer is depicted in Figure~\ref{fig:professionalphotographer}.

In our work, we have taken into consideration these two groups of characteristics and modeled our proposed solution accordingly. 
% We tuned the pipeline parameters using the images provided as \textit{train} images, while evaluating the results respectively on the two validation groups of images and the final test set.
Figure~\ref{fig:hist_pipe} shows the effect of each step of the pipeline on the image by showing the histogram of the example image alongside the output of each step. As can be seen, the pipeline has been designed in order to respect the image aesthetic described the professional photographer.

% https://nightimaging.org/expert-opinion-first-validation.html
% https://nightimaging.org/expert-opinion-second-validation.html
% https://nightimaging.org/expert-opinion-final-validation.html
\begin{comment}
commenti rilevanti dell esperto fotografo:
FIRST SUBMISSION
- Artifact-free.
- Overall fairly neutral colour balance with colourful small elements. If there is any colour cast, blue is more acceptable, while greens (from cyan to yellow-green) are by tradition unacceptable.
- Full tonal range, ie black point set.
- Unlit and weakly lit areas dark, ie histogram bias to left
-No clipping except for point light sources and speculars.
-Saturation not to reach 100\%, which reads as unrealistic. This is particularly important for night scenes, featuring both light sources and illuminated small areas against an overall dark background, which enhances brightness. Both the Hunt effect (saturation increases with brightness) help exaggerate these.
SECOND SUBMISSION
- With all the above in mind, the scene should not look like daytime. In most cases there should already be sufficient clues that it is night-time, but in some cases it might be desirable to lower the overall brightness. This borders on taste (see below).
- Overall quite colourful.

\end{comment}

% si intende i commmenti dell esperto fotogrago? cioè dire che ha detto questo e questo e quindi abbiamo cercato di seguire le sue linee guida sui risultati?

\subsection{Results and Discussion}

Our pipeline has been evaluated by comparing it with the other challenge solutions \cite{ntire2022} using Mean Opinion Score (MOS) obtained through vision comparison on the Yandex Toloka platform. Here every submission, consisting of 50 images of the final validation set, was included in 3250 comparisons. The results of the final leaderboard are reported in Table~\ref{tab:mos}. As shown, our pipeline won the fifth place of the challenge obtaining 1935 votes. Note that our solution received only 112 fewer votes than the second winning solution (i.e. about 5\% less votes) that uses different neural models for most of the operations in its pipeline \cite{ntire2022}. 
\begin{table}
    \centering
    \adjustbox{width=\columnwidth}{
    \begin{tabular}{|cccc|r|}
    \hline
         \textbf{Rank} & \textbf{Team} & \textbf{Score} & \textbf{Votes} & \textbf{Sign. Score} \\\hline\hline
        1 & xxx & 0.8009 & 2603 & 12\\\hline
        2 & xxx & 0.6298 & 2047 & 11\\\hline
        3 & xxx & 0.6089 & 1979 & 11\\\hline
        4 & xxx &	0.6045 & 1964 & 11\\\hline
        \textbf{5} & \textbf{xxx} & \textbf{0.5955} &	\textbf{1935} & \textbf{10}\\\hline
        6 & xxx & 0.5742 & 1866 & 9\\\hline
        7 & xxx & 0.4798 & 1559 & 6\\\hline
        8 & xxx & 0.4631 & 1505 & 6\\\hline
        9 &	xxx &	0.4411 & 1433 & 5\\\hline
        10 & xxx & 0.3965 & 1288 & 3\\\hline
        11 & xxx & 0.3683 & 1197 & 3\\\hline
        12 & Baseline & 0.2734 & 888 & 1 \\\hline
        13 & xxx & 0.0182 & 59 & 0\\\hline
    \end{tabular}}
    \caption{Final leaderboard of the NTIRE2022 Night Photography Rendering challenge. Every submission (50 images) was included in 3250 comparisons using the Yandex Toloka platform. Team names obscured for double blind review. Our team is highlighted in bold.}
    \label{tab:mos}
\end{table}
In Table~\ref{tab:mos} we also add a further column, named significance score. First of all for each method we compute the 95\% confidence interval of the Score using the Binomial test. The significance score for each solution corresponds to the number of solutions with respect to which it is statistically better or equivalent, i.e. the number of confidence intervals that are lower or overlap with the current one. The significance score highlights how the result achieved by the first solution is statistically better than all the others, while the solutions ranked from the second position to the fourth one are actually statistically equivalent and therefore rank in the second place. They are followed by our solution, that ranks in the third place and is statistically better than all the remaining solutions.   
%This suggests that traditional imaging pipelines can still compete with learning-based methods, which usually require a high number of training images and are characterized by high complexity. 
The results have been additionally evaluated by a professional photographer, who awarded our solution with the sixth place in the final leaderboard \cite{ntire2022}.

\begin{figure*} 
    \centering
    \begin{tabular}{cc}
        \textsc{Baseline pipeline} & \textsc{Our pipeline} \\
        \includegraphics[width=0.48\textwidth]{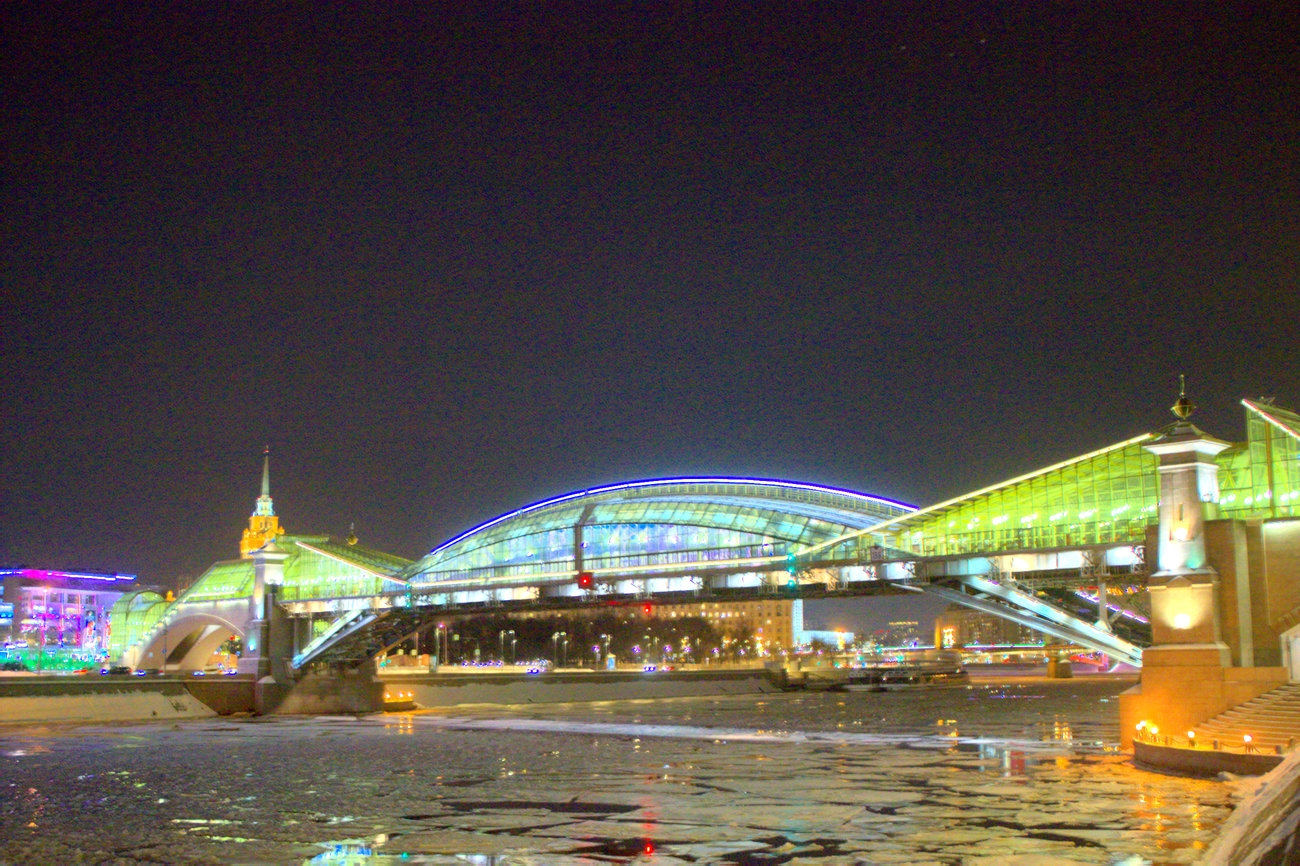} & \includegraphics[width=0.48\textwidth]{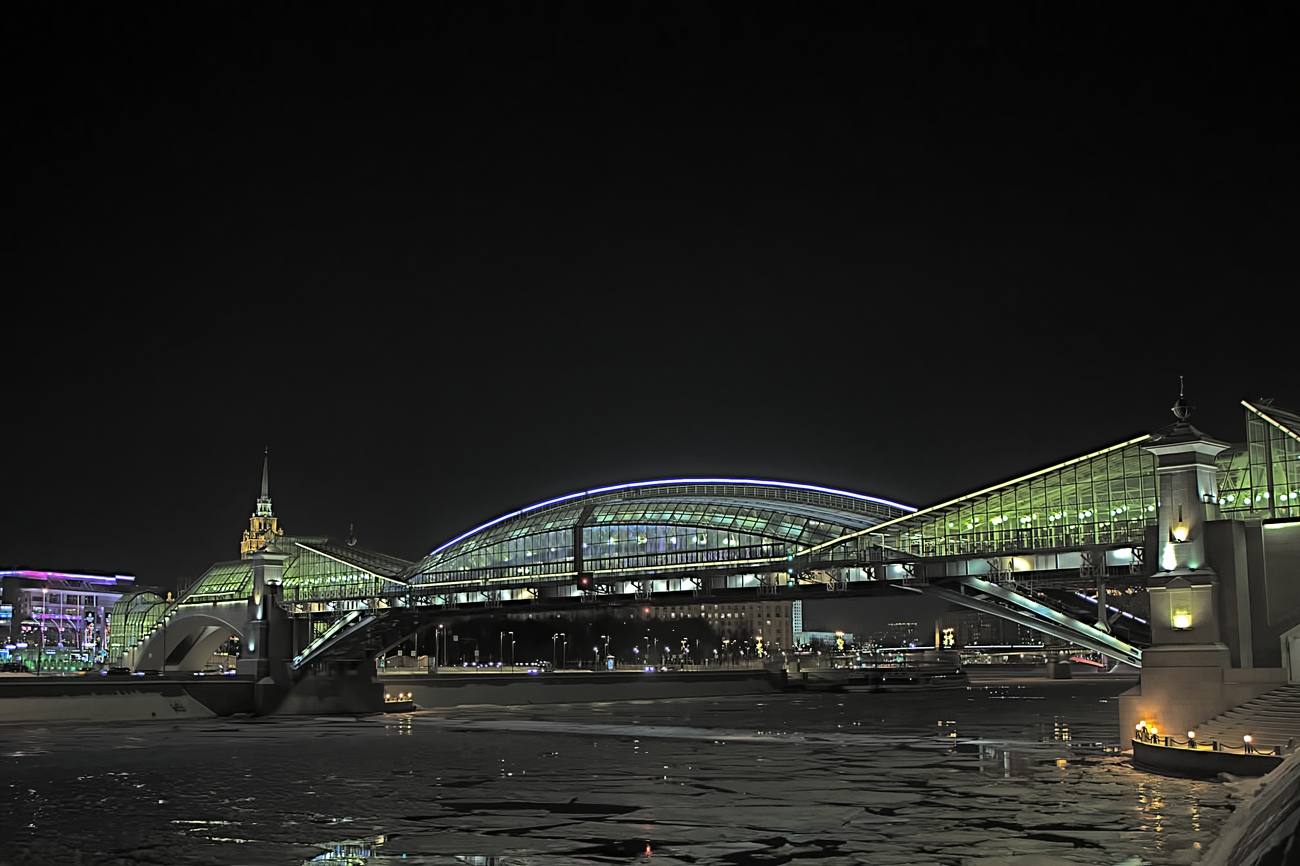} \\
        \includegraphics[width=0.48\textwidth]{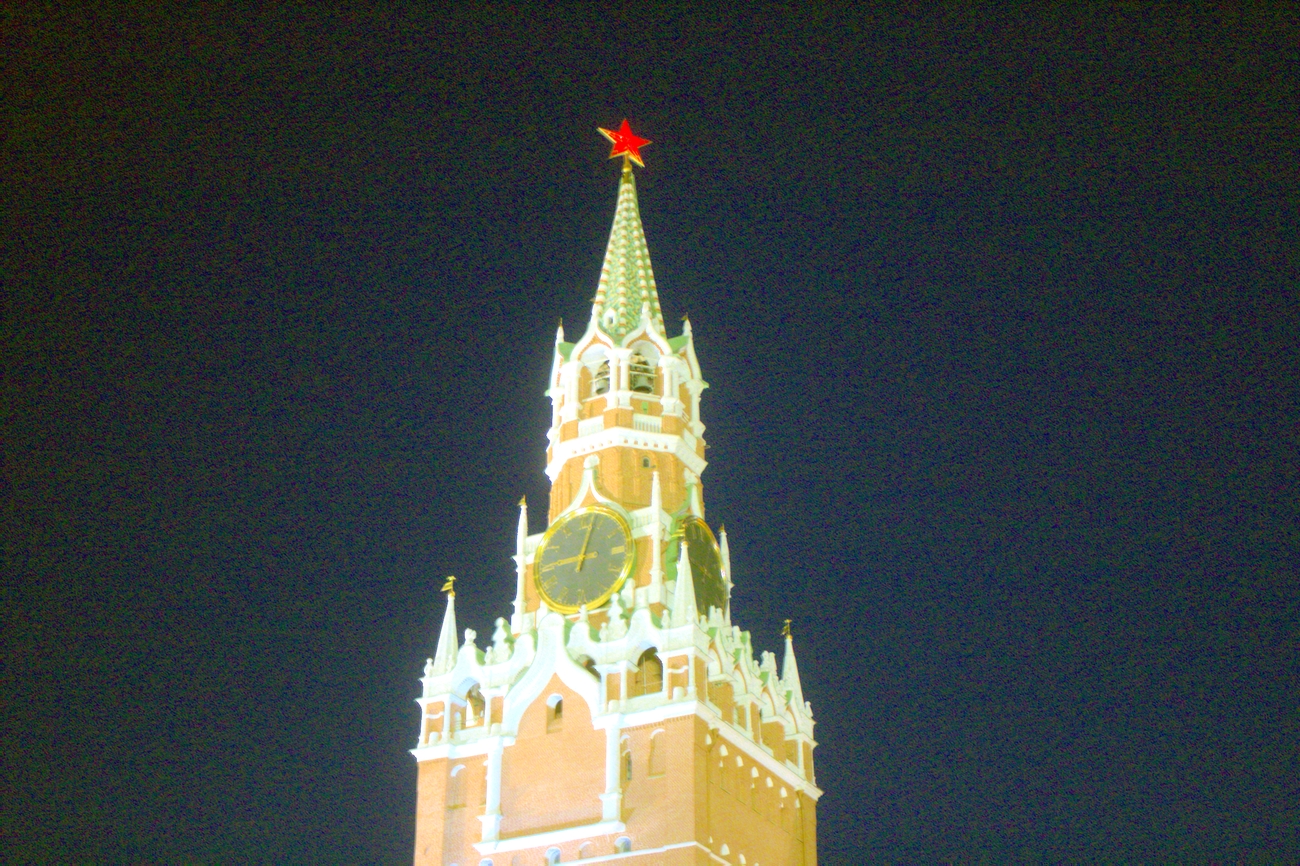} & \includegraphics[width=0.48\textwidth]{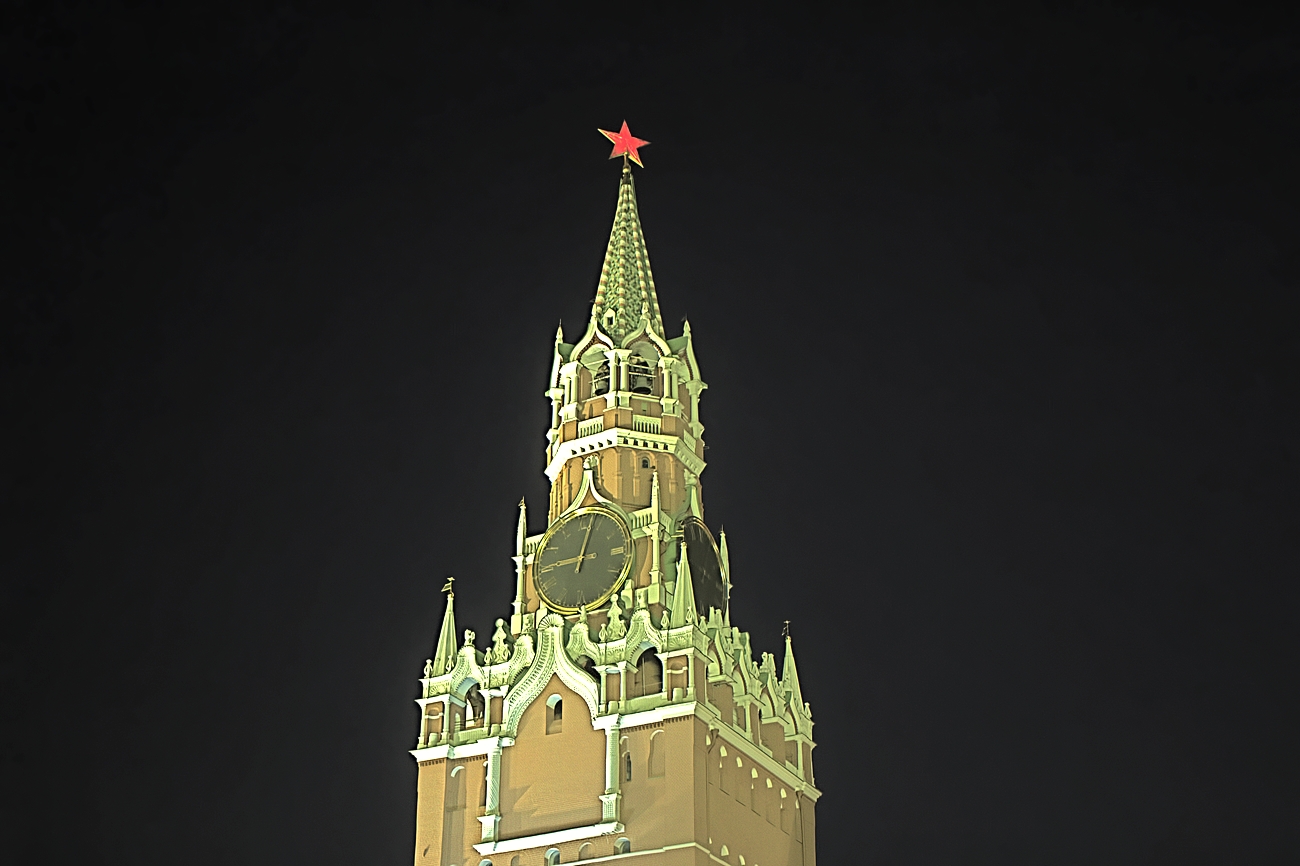} \\
        \includegraphics[width=0.48\textwidth]{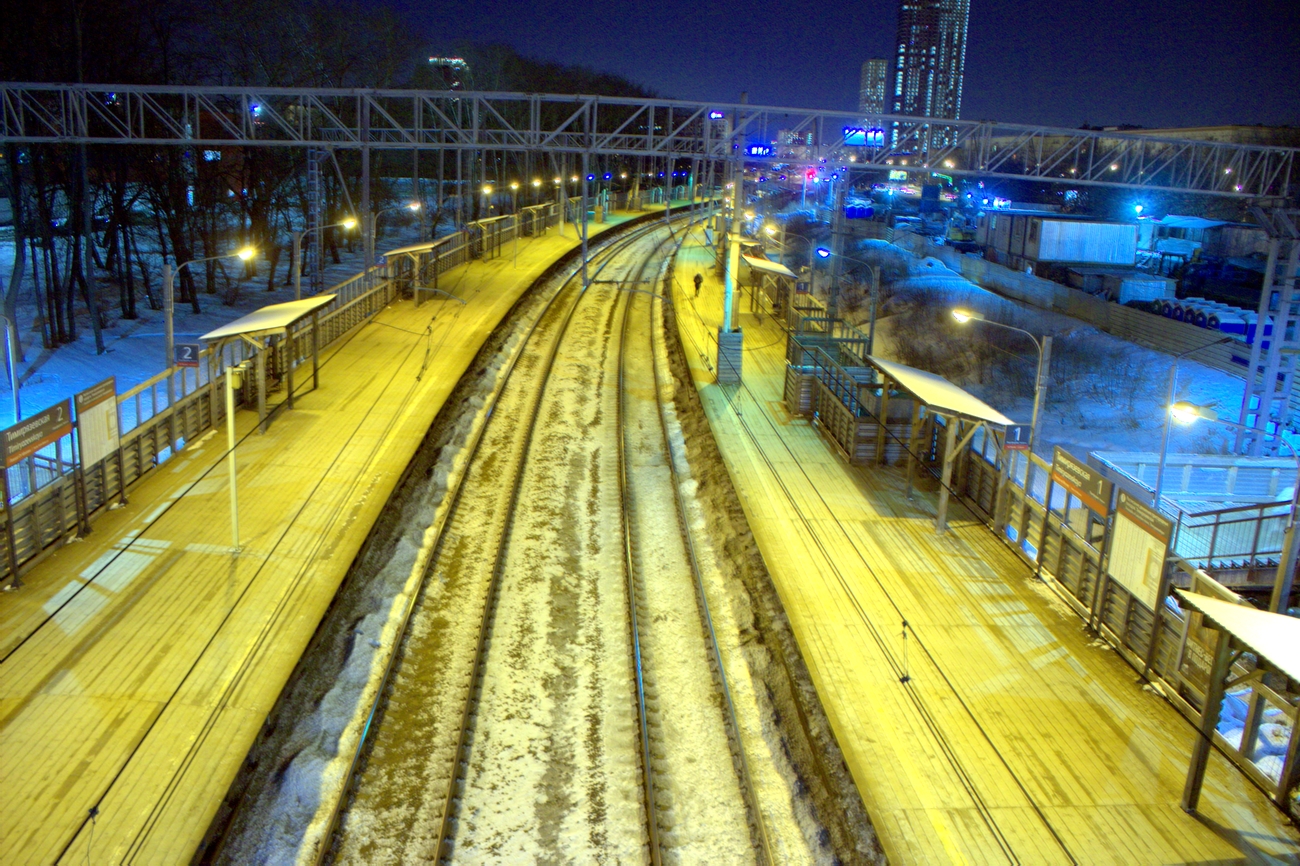} & \includegraphics[width=0.48\textwidth]{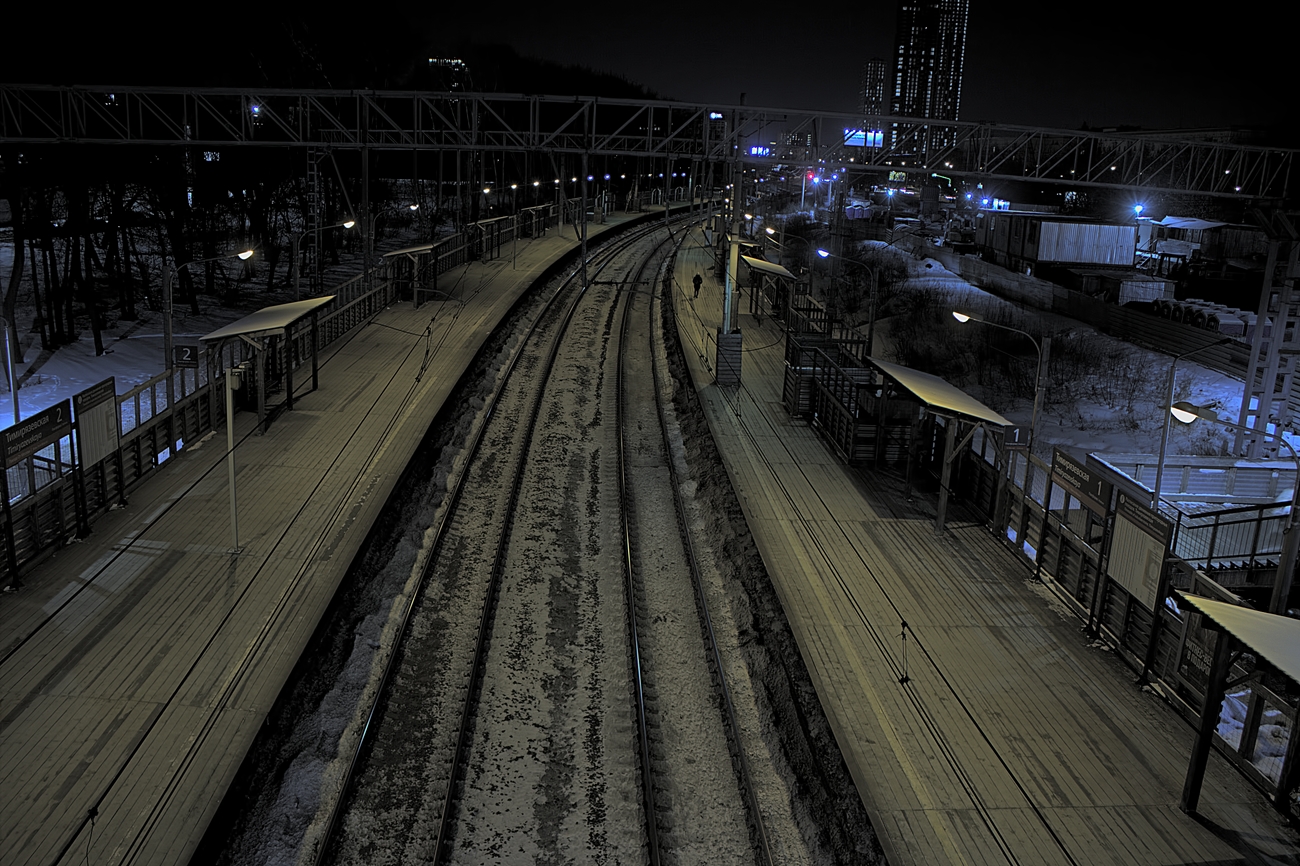} \\
    \end{tabular}
    \caption{Visual comparison between the results of the baseline pipeline \cite{baseline} of the NTIRE2022 Night Photography Rendering challenge (left) and the proposed pipeline (right). Our solution produces more ``visually pleasing'' results, reduces noise and color cast, and better maintains the mood of night photographs.}
    \label{fig:examples}
\end{figure*}

Some visual results of the proposed pipeline are shown in Figure \ref{fig:examples}. Here we also reported the same images corrected using the baseline pipeline \cite{baseline} provided by the challenge organizers for comparison. From the examples reported, we can observe how our solution can better remove noise, reduce color cast and preserve the mood typical of night scenes producing more ``pleasing'' results, as also confirmed by the score in the final leaderboard.

The proposed pipeline can process $3646\times5202$ RAW images and produce $866\times1300$ images in 61.34 seconds on a machine equipped with AMD Ryzen 9 3900X CPU and 16 GB of RAM. Note that the entire pipeline has been developed in Python without performing performance optimization in terms of execution time, which was out of the scope of the challenge. The detailed analysis of the running time of each step within the low-light specific part of our pipeline is documented in Table~\ref{tab:time}. Here the time is measured in seconds and computed as the average value of 10 runs. In particular, the bottleneck is the denoising step performed using BM3D that, alone, occupies the 90.64\% of the entire pipeline running time. However, the BM3D implementation we used is not optimized and a GPU-accelerated BM3D version \cite{bm3dgpu} can be used to considerably reduce the required time. Besides, one can easily replace BM3D with another denoising algorithm. For instance, the time of the denoising step performed by non-local means \cite{nonlocalmeans} reduces from 55.60 to 1.60 seconds, reducing the entire pipeline execution time from 61.34 to 6.96 seconds.
\begin{table}
    \centering
    \begin{tabular}{|cr|}
    \hline
         \textbf{Operation} & \textbf{Time (s)} \\\hline\hline
         \textbf{Preliminary steps} & \textbf{0.94} \\\hline\hline
         \textbf{Low-light specific steps} & \textbf{59.66} \\\hline 
         Local Contrast Correction \cite{mcc} & 0.51 \\%\hline
         Contrast + Saturation enhancement & 1.40 \\%\hline
         Black point + Gamma correction & 1.74 \\%\hline
         Sharpening & 0.17 \\%\hline
         %To uint8 + Resize & ? \\\hline
         BM3D \cite{bm3d} denoising & 55.60 \\%\hline
         GI \cite{GI} white balancing & 0.24 \\\hline\hline
         %Orientation fix & ? \\\hline 
         \textbf{Entire pipeline} & \textbf{61.34} \\\hline
    \end{tabular}
    \caption{Running time of each low-light specific step in our pipeline, expressed in seconds. The reported values are the average of 10 runs on a machine equipped with AMD Ryzen 9 3900X CPU and 16 GB of RAM. The total time of the pipeline also includes few minor steps (e.g., orientation fix) whose absence justifies the difference between the entire pipeline execution time and the summation of the single reported steps.}
    \label{tab:time}
\end{table}

Figure \ref{fig:hist_pipe} shows intermediate images and histograms of the proposed pipeline after each step. The application of LCC \cite{mcc} improves the local contrast but centers the image histogram and reduces the overall saturation, hence the contrast and saturation enhancement step is necessary to correct this behaviour. Yet, the obtained histogram is still biased towards the center of the dynamic range, and a black level adjustment is fundamental to restore the natural anesthetic of the image.
Here a gamma correction can increase the overall brightness. Since BM3D \cite{bm3d} effectively removes noise but also part of the details, a preliminary sharpening operation that strengthens high frequencies helps preventing this problem.

\section{Conclusions}
\label{sec:conclusions}
We have proposed a low-complexity handcrafted camera pipeline for the generation of visually-pleasing night photographs, as part of the NTIRE2022 Night Photography Rendering challenge.
Our solution depends from a small number of free parameters, which we empirically set according to our personal preference.
% Possibly we could have obtained better results by having experienced photographers help us in the choice of parameters, but for us winning the challenge was not the most important thing, as much as demonstrating that traditional imaging pipes can compete with modern deep learning-based methods. 
The final leaderboard results demonstrate that traditional imaging pipelines can compete with modern deep learning-based methods.
Furthermore, our approach has, in our view, significant advantages:
\begin{itemize}
    \item[-] It does not require large training sets.
%    \item[-] It is easily adaptable to different sensors.
    \item[-] It is easily adaptable for different types of users.
    \item[-] It is explainable.
    \item[-] It is computationally efficient.
\end{itemize}

The optimal parameters could be easily found with optimization methods if one had a suitable training set (whose cardinality however should not be so high as in the case of neural networks), which we consider an interesting research direction for future development, tuning the pipeline to different users and sensors.
Another promising research direction is that of exploiting saliency \cite{bianco2020neural} to perform a spatially varying enhancement, or to exploit no reference image aesthetic metrics (e.g. \cite{bianco2016predicting}) to drive model parameters selection.

%%%%%%%%% REFERENCES
{\small
\bibliographystyle{ieee_fullname}
\bibliography{egbib}
}

\end{document}